\documentclass[sigconf]{acmart}
\AtBeginDocument{%
  }

\setcopyright{acmlicensed}
\copyrightyear{2026}
\acmYear{2026}
\acmDOI{XXXXXXX.XXXXXXX}

\acmISBN{978-x-xxxx-xxxx-X/YYYY/MM}

\usepackage{array}
\usepackage{booktabs}
\usepackage{multirow}
\usepackage{multicol}

\usepackage{pifont}

\usepackage{CJKutf8}

\usepackage{float}
\usepackage{placeins}

\definecolor{purple}{RGB}{148, 39, 212}
\definecolor{lightRed}{RGB}{253, 190, 172}
\definecolor{lightBlue}{RGB}{172, 216, 253}

\newcolumntype{C}[1]{>{\centering\arraybackslash}p{#1}}

\newcolumntype{M}[1]{>{\centering\arraybackslash}m{#1}}

\newcolumntype{L}[1]{>{\raggedright\arraybackslash}p{#1}}

\begin{document}

\title{
RELATE: A Reinforcement Learning-Enhanced LLM Framework
for Advertising Text Generation
}

\author{Jinfang Wang}
\authornote{Both authors contributed equally to this research.}
\email{wangjinfang@baidu.com}
\affiliation{
  \institution{Baidu Inc.}
  \city{Beijing}
  \country{China}
}

\author{Jiajie Liu}
\authornotemark[1]
\email{liujiajie01@baidu.com}
\affiliation{
  \institution{Baidu Inc.}
  \city{Beijing}
  \country{China}
}

\author{Jianwei Wu}
\email{wujianwei@baidu.com}
\affiliation{
  \institution{Baidu Inc.}
  \city{Beijing}
  \country{China}
}

\author{Ziqin Luo}
\email{luoziqin@baidu.com}
\affiliation{
  \institution{Baidu Inc.}
  \city{Beijing}
  \country{China}
}

\author{Zhen Chen}
\email{chenzhen19@baidu.com}
\affiliation{
  \institution{Baidu Inc.}
  \city{Beijing}
  \country{China}
}

\author{Chunlei Li}
\email{lichunlei02@baidu.com}
\affiliation{
  \institution{Baidu Inc.}
  \city{Beijing}
  \country{China}
}

\author{Biao Han}
\email{hanbiao@baidu.com}
\affiliation{
  \institution{Baidu Inc.}
  \city{Beijing}
  \country{China}
}

\author{Tao Deng}
\email{dengtao02@baidu.com}
\affiliation{
  \institution{Baidu Inc.}
  \city{Beijing}
  \country{China}
}

\author{Yi Li}
\email{liyi01@baidu.com}
\affiliation{
  \institution{Baidu Inc.}
  \city{Beijing}
  \country{China}
}

\author{Shuanglong Li}
\email{lishuanglong@baidu.com}
\affiliation{
  \institution{Baidu Inc.}
  \city{Beijing}
  \country{China}
}

\author{Lin Liu}
\email{liulin03@baidu.com}
\affiliation{
  \institution{Baidu Inc.}
  \city{Beijing}
  \country{China}
}

\renewcommand{\shortauthors}{Wang et al.}



\begin{abstract}
In online advertising, advertising text plays a critical role in attracting user engagement and driving advertiser value. Existing industrial systems typically follow a two-stage paradigm, where candidate texts are first generated and subsequently aligned with online performance metrics such as  click-through rate(CTR). This separation often leads to misaligned optimization objectives and low funnel efficiency, limiting global optimality.

To address these limitations, we propose \textbf{RELATE}, a reinforcement learning-based end-to-end framework that unifies generation and objective alignment within a single model. Instead of decoupling text generation from  downstream metric alignment, RELATE integrates performance and compliance objectives directly into the generation process via policy learning. To better capture ultimate advertiser value beyond click-level signals, We incorporate conversion-oriented metrics into the objective and jointly model them with compliance constraints as multi-dimensional rewards, enabling the model to generate high-quality ad texts that improve conversion performance under policy constraints.

Extensive experiments on large-scale industrial datasets demonstrate that RELATE consistently outperforms  baselines. Furthermore, online deployment on a production advertising platform yields statistically significant improvements in click-through conversion rate(CTCVR) under strict policy constraints, validating the robustness and real-world effectiveness of the proposed framework.

\end{abstract}







\maketitle
\begin{figure}[h]
\centering 
\captionsetup{aboveskip=0pt, belowskip=0pt} 
\includegraphics[width=1.0\linewidth,page=1]{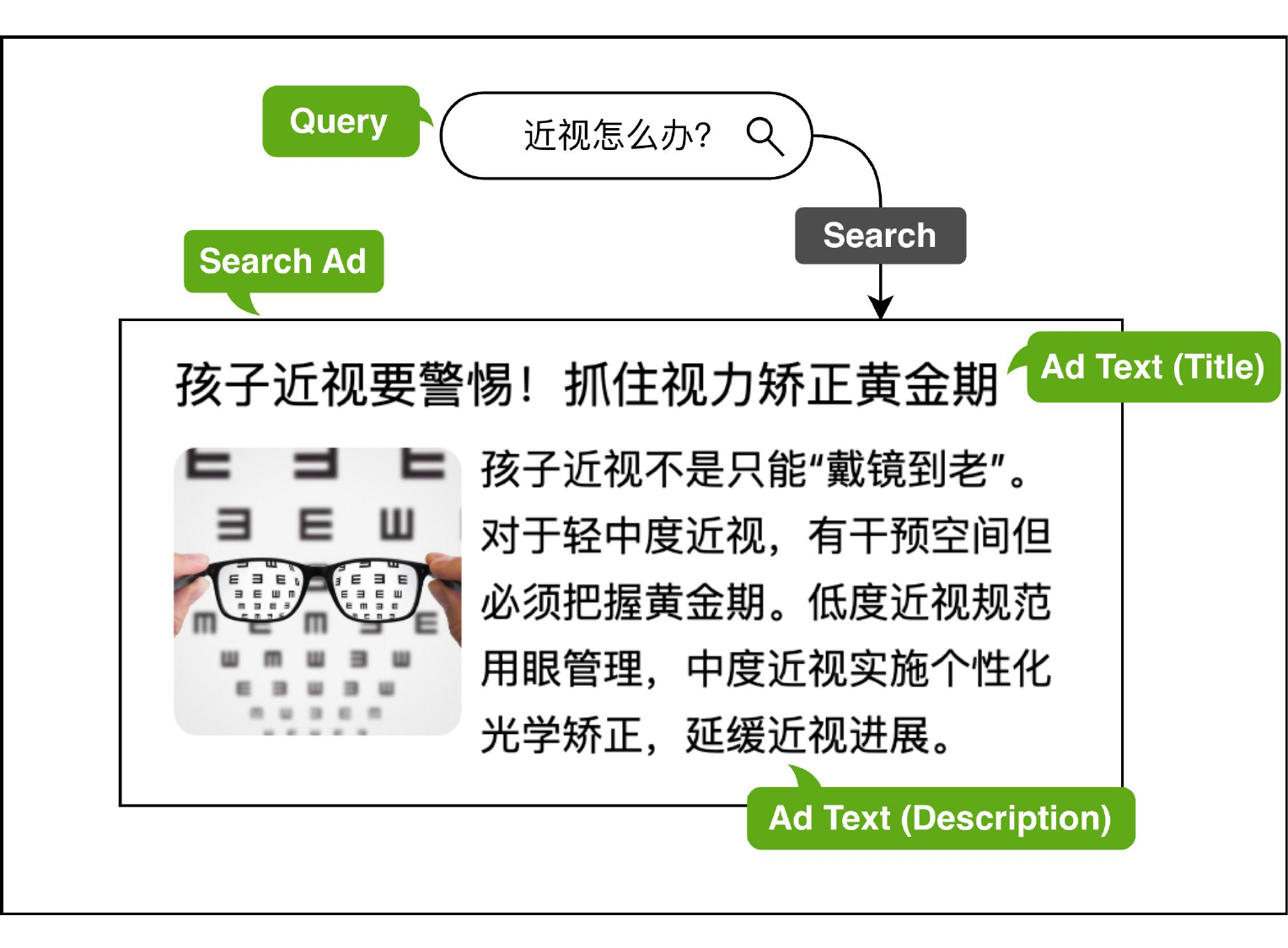}
\vspace{0.01cm}
\caption{
An ad text illustration of online advertisement
}
\label{fig:product}
\end{figure}

\section{Introduction}
In online advertising systems, advertising text serves as a critical bridge between user intent and commercial services (shown in Figure~\ref{fig:product}), playing a crucial role in attracting user attention and improving CTCVR. 
The ability to automatically generate advertising texts that are both high-quality and conversion-effective, while remaining suitable for long-term deployment, therefore represents a central technical challenge for modern advertising platforms.

Existing advertising text generation methods can be broadly categorized into offline-metric-driven and online-metric-driven approaches. 
Offline-metric-driven methods primarily focus on intrinsic text quality measures such as fluency, diversity, or stylistic consistency, typically based on large language models (LLMs) and relying on prompt engineering or supervised fine-tuning (SFT) for generation. 
While these methods perform well on offline automated evaluations such as quality benchmarks~\cite{zhang2024adtec} that assess fluency and diversity, they lack direct alignment with real commercial outcomes such as CTR or conversion rates(CVR), limiting their online effectiveness.
Compared to offline-quality-driven approaches, online-metric-driven methods aim to optimize online performance objectives such as CTR~\cite{zhou2023creater,chen2025ctr}.
Most existing approaches follow a two-stage paradigm, where diverse candidate advertising texts are first generated using rules or generative models, and then aligned with online feedback signals (e.g., click logs) via contrastive learning or preference optimization techniques such as DPO~\cite{rafailov2023dpo}.
While these methods are more closely aligned with online performance goals, the decoupling between generation and optimization often leads to objective inconsistency, low funnel efficiency, and increased system complexity, which fundamentally limits the upper bound of achievable performance.

Advertising text generation can be formulated as a constrained optimization problem, where alignment with the advertiser’s business objectives and correctness of advertising text serve as constraints, and online performance metrics such as CTR or CTCVR serve as the optimization objective. 
Ideally, a unified model should jointly satisfy these constraints while directly optimizing online performance. However, achieving such a unified solution in practice remains challenging.

Current industrial approaches for advertising text generation primarily rely on natural language generation models trained via maximum likelihood estimation (MLE). 
Such models optimize the likelihood of historical training data, which often leads to a misalignment with practical online objectives(CTR or CTCVR), as well as constraints like relevance to the advertiser’s business. 
In addition, advertising text generation faces the challenge of text fatigue~\cite{nlg_ad_survey2026}, where repeatedly exposing the same advertising text to a user can reduce engagement and degrade CTR and CTCVR. 
This problem is further exacerbated by the limited supply of diverse advertising texts. Consequently, designing models that can jointly align optimization objectives, maintain relevance, and encourage diverse generation to mitigate fatigue remains a major challenge in real-world systems.

In this paper, we propose \textbf{RELATE}, a \textbf{RE}inforcement learning-enhanced \textbf{L}LM framework for \textbf{A}dvertising \textbf{TE}xt generation. 
It aims to address the key challenges in advertising text generation. 
Specifically, to resolve the misalignment between the intrinsic training objective of language models and advertising-specific optimization goals, we construct dedicated rewards for both constraint satisfaction and business objectives, and directly integrate them into the generation process via reinforcement learning, enabling end-to-end objective alignment. 
To further encourage effective exploration and discover higher-quality advertising texts, we design diversity-aware rewards and incorporate them into reinforcement learning training to explicitly promote diverse generation. 
Moreover, to tackle the reward sparsity and delayed feedback issues commonly observed in large language model reinforcement learning, we propose a task-specific credit assignment strategy tailored to advertising generation, significantly improving training efficiency and convergence behavior.

Since deployment in a large-scale production environment, our approach has achieved a relative increase of 9.19 \% in CTCVR compared to the production baseline, further validating its practical effectiveness and scalability in real-world advertising systems. The main contributions of our work can be summarized as follows:
\begin{enumerate} 
\item We formulate advertising text generation as a constrained optimization problem and propose a reinforcement-learning-enhanced, end-to-end large language model framework that directly optimizes online  performance objectives, departing from the conventional two-stage generate-and-align paradigm.
\item We design a unified reward modeling framework that jointly captures creative quality, diversity and conversion rate, enabling effective reinforcement learning for advertising creative generation.
\item We propose a task-specific credit assignment strategy to address reward sparsity and delayed feedback, substantially improving training efficiency and convergence behavior.
\item Extensive offline and online experiments demonstrate consistent improvements on relevance-related metrics and online performance metrics such as CTCVR.
\end{enumerate}
The remainder of this paper is organized as follows. We first review related industrial and academic work on advertising text generation. We then formally define the dynamic advertising text generation task and present the proposed reinforcement-learning-based end-to-end modeling framework in detail. Next, we describe the experimental settings and report experimental results along with detailed analyses. Finally, we conclude the paper.

\section{Related Work}
\textbf{Advertising Text Generation.}
Early studies on advertising text generation were mainly based on template-driven approaches~\cite{bartz2008natural,thomaidou2013grammads}.
These methods generate advertising texts at scale by filling predefined, manually designed templates with keywords, offering advantages such as simplicity and strong controllability. 
However, their expressive capacity is inherently constrained by the size of the template set and handcrafted rules, making it difficult to cover complex and diverse linguistic patterns. 
As a result, both text diversity and long-term performance improvements are limited.
With the rapid development of pretrained language models and large-scale generative models, researchers have increasingly explored neural generation–based approaches for advertising text generation~\cite{jin2023towards,zhang2021chase,youngmann2021algorithmic,vasudevan2025llm}. 
CHASE~\cite{zhang2021chase} enhances the attractiveness of generated advertising text by integrating keywords, commonsense knowledge, and marketing corpora into language model generation. 
Youngmann et al.~\cite{youngmann2021algorithmic} proposes a representative two-stage framework, where a generative model first produces candidate ad texts based on URLs or landing pages, followed by a transformation model that rewrites and optimizes these candidates. 
More recently, Vasudevan et al.~\cite{vasudevan2025llm} introduces a multi-stage Generate–Evaluate–Refine framework built upon large language models to address the multiple complex constraints involved in e-commerce banner ad generation.
Despite notable progress in fluency and expressiveness, these approaches are still largely driven by offline language quality metrics or heuristic objectives, resulting in a disconnect between the generation process and real-world online advertising objectives.

\textbf{Diversity-aware Text Generation.}
In advertising delivery scenarios, repeatedly exposing highly similar advertising texts to the same user or similar user groups often leads to text fatigue,which can reduce user attention.Consequently, diversity-aware generation has become a critical research direction in advertising text production.
Existing studies have explored diversity enhancement from multiple perspectives. 
PLANNER~\cite{zhang2023planner} proposes a hybrid model that combines latent semantic diffusion with autoregressive decoding to generate long-form text that is both high-quality and diverse. 
CTOP~\cite{chen2025ctr} introduces a chain-of-thought–based in-context learning framework that enhances generation diversity by incorporating reasoning-augmented examples via retrieval-augmented generation (RAG). 
In contrast, POEM~\cite{cao2025perspective} explicitly incorporates diversity optimization objectives into natural language generation by constructing diversity preference data and aligning models using DPO, while further introducing a semantic entropy loss to encourage diversity at the output distribution level.
However, most of these diversity-oriented methods primarily focus on linguistic or semantic variation at the text level. They rarely incorporate online feedback signals from advertising systems for joint optimization, making it difficult to ensure alignment between diversity exploration and business objectives.

\textbf{CTR-driven Ad Text Generation.}
The ultimate goal of advertising text generation is to improve online performance metrics, such as CTR.
Accordingly, a growing body of recent work has explored reinforcement learning or preference alignment mechanisms to directly optimize non-differentiable business metrics.
Early work~\cite{hughes2019generating} adopts a hybrid training paradigm that combines supervised learning and reinforcement learning: a seq2seq model is first trained to imitate human-written advertisements, and then optimized using CTR-related signals as rewards during reinforcement learning. 
Building upon this, UMPG~\cite{wang2021reinforcing} proposes a three-stage pretraining–fine-tuning–reinforcement learning framework, where CTR, writing quality, and faithfulness rewards are jointly incorporated during the reinforcement learning stage to optimize online CTR under basic quality constraints. 
CREATER~\cite{wei2022creater} integrates CTR signals into model optimization through contrastive fine-tuning, leveraging preference data mined from online A/B experiments to guide the generation of higher-CTR advertising texts. 
CTOP~\cite{chen2025ctr} further explores preference alignment methods based on CTR gains with confidence-weighted signals to improve the stability of generation outcomes.
While these approaches partially bridge the gap between generative models and business objectives, most still follow multi-stage generation–optimization paradigms, which suffer from low funnel efficiency and inconsistent objectives across stages. 
Moreover, quality constraints, diversity exploration, and CTR optimization are often modeled separately, lacking a unified end-to-end optimization framework.

In summary, existing research on advertising text generation has made substantial progress in terms of generation quality, diversity modeling, and CTR-driven optimization. Nevertheless, several key limitations remain:
1) multi-stage paradigms lead to fragmented optimization objectives;
2) diversity modeling is insufficiently integrated with online performance metrics.
Motivated by these limitations, this work proposes an end-to-end, unified, online-metric-driven framework for advertising text generation.

\section{Methodology}
In this section, we first formally define the advertising text generation task
~(Section~\ref{sec:task_definition}). 
We then present an overview of the proposed end-to-end unified framework~(Section~\ref{sec:framework}), followed by detailed descriptions of the reward function design~(Section~\ref{sec:reward_design}) and the credit assignment mechanism together with the reinforcement learning algorithm~(Section~\ref{sec:credit_assign_reinforcement_learning}).

\begin{figure*}[h]
\centering 
\captionsetup{aboveskip=0pt, belowskip=0pt} 
\includegraphics[width=1.0\linewidth,page=1]{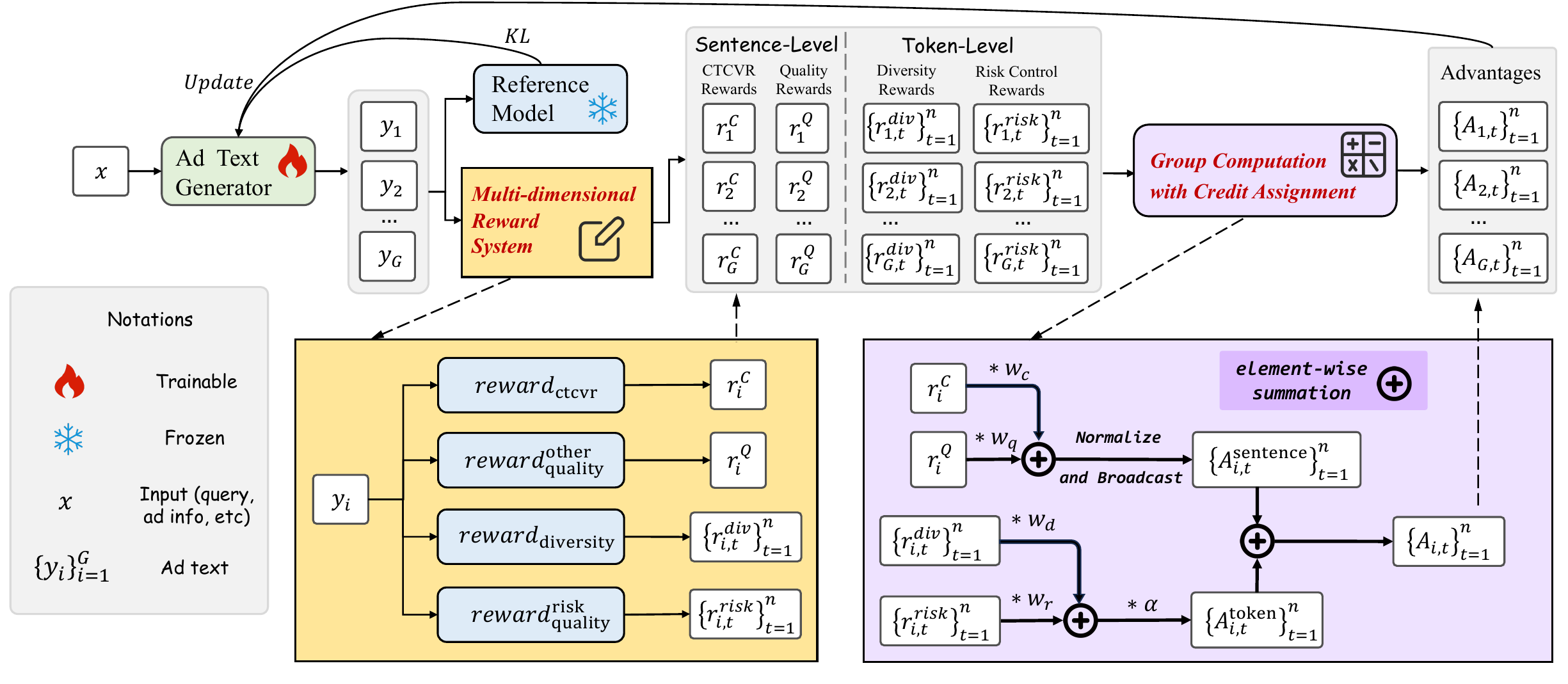}
\caption{
Workflow overview of RELATE. 
Built upon GPRO~\cite{shao2024grpo}, RELATE features a \textit{Multi-dimensional Reward System} and a \textit{Group Computation with Credit Assignment}. 
The former integrates rewards from multiple reward dimensions, including CTCVR Reward, Quality Reward, and Diversity Reward. 
The latter introduces credit assignment during group computation to derive differentiated, token-level advantages.
}
\label{fig:teaser}
\end{figure*}

\subsection{Task Definition}
\label{sec:task_definition}
We formulate advertising text generation as a conditional generation problem that aims to maximize advertising conversion while simultaneously satisfying quality and diversity constraints.
Given an input $x$, which consists of user intent expressions (queries) and advertising-related information (e.g., bidwords, landing page content), the goal is to learn a generation policy $\pi_\theta$ that produces advertising text $y$. 
The generated texts should satisfy quality and diversity requirements while maximizing online performance metrics such as CTCVR.
In its idealized form, the problem can be expressed as the constrained optimization objective
\begin{equation}
\label{eq:idea_optimization_goal}
\begin{matrix}
\max\quad & \text{ctcvr}(y=\pi_\theta(x))\\
\text{s.t} & \text{quality}(y) \geq \tau_q \\
 & \text{diversity}(y) \geq \tau_d \\
\end{matrix},
\end{equation}

where $\tau_q$ and $\tau_d$ denote the minimum acceptable thresholds for content quality and diversity, respectively.

\subsection{Framework}
\label{sec:framework}
Directly solving problem~(\ref{eq:idea_optimization_goal}) is challenging in large-scale generation settings. 
Therefore, unlike traditional approaches that follow a generate-first, then align via multi-stage ranking or filtering paradigm~\cite{youngmann2021algorithmic,vasudevan2025llm,chen2025ctr,wang2021reinforcing}, we propose an end-to-end unified advertising text generation framework that directly aligns business objectives with the generation process through reinforcement learning. 
The overall workflow is illustrated in Figure~\ref{fig:teaser}.
Specifically, we unify conversion objectives and various generation constraints into a single scalar reward function
\begin{equation}
    \label{eq:total_reward}
    reward(x,y)=f(\text{ctcvr}(x,y), \text{quality}(y), \text{diversity}(y)),
\end{equation}
where reward components \text{ctcvr}, \text{quality} and \text{diversity} correspond to online conversion performance, semantic and compliance-related quality, and text diversity respectively.
Under this formulation, advertising text generation is optimized via the following reinforcement learning objective
\begin{equation}
    \label{eq:reward_optimization_goal}
    \max_{\pi_{\theta}} \mathbb{E}_{x \sim \mathcal{D},\, y \sim \pi_{\theta}(y \mid x)}\!\left[reward(x, y)\right].
\end{equation}
Through optimizing (\ref{eq:reward_optimization_goal}), quality and diversity requirements, originally defined as hard constraints, are softened and incorporated directly into the generation stage, enabling end-to-end objective alignment.

This unified framework offers two key advantages:
1) \textbf{funnel compression and objective consistency.} 
By integrating generation and reward optimization, performance degradation caused by inconsistent objectives across multiple stages is avoided.
2) \textbf{Direct align advertising text with business metrics.} 
Online conversion metrics serve as the primary optimization target, providing a direct guidance for continuous reinforcement learning–based improvement and increasing the advertising system’s performance ceiling.

\subsection{Reward Design}
\label{sec:reward_design}
This section introduces the design of the multi-dimensional reward framework used in RELATE. 
The reward function integrates quality, diversity, and conversion objectives, providing structured guidance for reinforcement learning-based advertising text generation under practical product and business constraints.

\subsubsection{Quality Reward}
The quality reward is designed to enforce essential product-level constraints in advertising generation, ensuring that model outputs are usable, compliant, and aligned with the advertising context. 
It provides foundational guidance during training and prevents the model from optimizing business objectives at the expense of basic content quality.
Specifically, it consists of the following components:
\begin{itemize}
    \item \textbf{Length Reward.} 
    A target token-length interval is defined for generated advertising texts. Outputs within the interval receive positive rewards, while overly short or long generations are penalized.
    \item \textbf{Format Reward.} 
    Structural constraints required by downstream serving or review pipelines are enforced, rewarding texts that satisfy predefined formatting rules or template requirements.
    \item \textbf{Relevance Reward.}
    Semantic alignment between the generated advertising text and the advertising context is evaluated, encouraging content that is consistent with the promoted product or service.
    \item \textbf{Correctness Reward.}
    Outputs containing factual errors or violations of common knowledge are penalized, improving content reliability.
    \item \textbf{Risk Control Reward.}
    Potentially risky or non-compliant content, such as exaggerated or misleading claims, is discouraged based on historical violation patterns and semantic judgment.
\end{itemize}
Together, these rewards establish a minimal yet sufficient quality constraint set, providing a stable foundation for optimizing business-oriented objectives.

\subsubsection{Diversity Reward}
Advertising text diversity is a fundamental requirement in advertising text generation, as it directly affects long-term user perception and the stability of advertising text distributions under large-scale deployment. 
Beyond optimizing the quality and performance of individual text, effective systems must preserve diversity at the distributional level.
During reinforcement learning optimization, models often over-exploit a limited set of high-reward expression patterns, converging to repetitive words, phrases, or sentence templates. 
This phenomenon, commonly referred to as mode collapse, may increase short-term expected reward but substantially degrades long-term effectiveness.
To address this issue, we explicitly incorporate diversity constraints into reward modeling. 
Diversity is formalized as the suppression of template-like behavior at the lexical and phrasal levels through an attributable diversity metric. 
Concretely, we compute the distribution of generated n-grams($n \in [2,10]$) and identify a set of high-frequency n-grams $\mathcal{G}_{high}$. 
For a generated advertising text $y$, the diversity penalty is defined as
\begin{equation}
    \label{eq:reward_diversity}
    reward_\text{diversity} = 
    -\sum_{g \in \text{n-grams}(y)} \mathbb{I}(g \in \mathcal{G}_{high}) \cdot \Delta d ,
\end{equation}
where $\mathbb{I}(\cdot)$ is an indicator function and $\Delta d$ controls penalty strength.
This design enables precise attribution of diversity penalties to specific high-frequency expressions, effectively mitigating template convergence within batches without compromising semantic coherence.

\subsubsection{CTCVR Reward}
The conversion reward aligns reinforcement learning optimization with real-world business objectives by directly incorporating predicted conversion performance. 
We adopt CTCVR as the primary target, thereby bridging offline training signals and online advertising effectiveness.
Specifically, user queries, ad metadata, and advertising texts are encoded into a shared semantic space using \texttt{Qwen3-Embedding-0.6B}~\cite{zhang2025qwen3embedding}. 
Based on these representations, a Shared-Bottom multi-task learning framework is employed to jointly model CTR and CTCVR, following the widely adopted entire-space modeling paradigm in industrial advertising systems~\cite{ma2018entire}, producing predicted CTCVR estimates for generated texts. 
The estimates are used as reward signals during reinforcement learning.

\subsection{Credit Assignment and Reinforcement Learning}
\label{sec:credit_assign_reinforcement_learning}
The reinforcement learning community has long focused on the credit assignment problem\cite{segmentpo2024, capo2025}, how to properly attribute final rewards to individual actions or decisions, which is considered a key challenge for improving policy learning efficiency and performance\cite{grpolambda2024, stop_summation2024}. In this work, we propose a granularity-aware credit assignment mechanism tailored to the characteristics of advertising text generation. The mechanism distinguishes between sentence-level and token-level rewards, and further enables fine-grained attribution for token-level rewards, enhancing overall model performance.

\subsubsection{Granularity-Aware Credit Assignment}

In advertising text generation, the reward signals relied upon by reinforcement learning exhibit significant heterogeneity in their natural attribution granularity. Some rewards can be directly attributed to individual token generation behaviors, while others can only be reliably evaluated after a complete response is generated. Specifically:
\begin{itemize}
    \item \textbf{Token-level rewards}, which are triggered by individual tokens, e.g., penalties arising from blacklisted words or high-frequency patterns in risk control or diversity constraints.
    \item \textbf{Sentence-level rewards}, which depend on the semantic and overall quality of the complete response and cannot be reliably decomposed to individual tokens, e.g., predicted conversion rate, relevance, and correctness rewards.
\end{itemize}
Based on this categorization, we propose a credit assignment strategy consistent with the reward generation mechanism, whose core principle is to allocate credit according to the reward’s natural granularity without introducing additional attribution assumptions.
For token-level rewards, we define the reward function directly at the token level. For instance, for risk control and diversity constraints, the token-level reward is formulated as
\begin{equation}
r_t^{\text{token}} =- \lambda_1 \mathbb{I}(T_t \in \mathcal{V}^{\text{black}}) - \lambda_2 \mathbb{I}(T_t \in \mathcal{G}_{high}),
\end{equation}
where $\mathcal{V}^{\text{black}}$ denotes the pre-defined blacklist of risky words and $\mathcal{G}_{high}$denotes the set of high-frequency n-gram patterns exceeding a threshold, $\mathbb{I}(\cdot)$ is the indicator function, and $\lambda_1,\lambda_2$ are the corresponding penalty coefficients. This design enables credit to be assigned at the computation stage, directly affecting the tokens that trigger the constraints.
For sentence-level rewards, due to the lack of reliable token-level attribution, we retain them as response-level reward signals for unified modeling and optimization in the subsequent reinforcement learning alignment phase.

\subsubsection{Reinforcement Learning Alignment}
In this work, we adopt GRPO (Generalized Reward-based Policy Optimization)~\cite{shao2024grpo} for reinforcement learning alignment
\begin{align}
    \label{eq:rl_final_objective}
    \mathcal{L}(\theta) =  \frac{1}{G} \sum_{i=1}^{G} \sum_{t=1}^{n} [ &\min \left( \rho_{i,t} A_{i,t}, \text{clip}(\rho_{i,t}, 1-\epsilon, 1+\epsilon) A_{i,t} \right) \notag \\
    &- \beta D_{KL}(\pi_\theta \| \pi_{\text{ref}})],
\end{align}
where $\rho_{i,t} = \frac{\pi_\theta(T_{i,t} | x, T_{i,<t})}{\pi_{\theta_{\text{old}}}(T_{i,t} | x, T_{i,<t})}$ denotes the importance sampling ratio. The KL regularization constrains the policy from deviating excessively from the pretrained language model, thereby maintaining fluency.
Building upon the credit assignment method introduced in Section 3.4.1, the Advantage$A_{i,t}$ is computed as follows:

\textbf{Token-determined Advantage.}
For token-level rewards that depend on individual tokens, we construct token-determined advantages directly from token-level feedback
\begin{equation}
A_{i,t}^{\text{token}} =\alpha * r_{i,t}^{\text{token}}, \quad t=1, \dots, n
\end{equation}
where $\alpha$ controls the relative importance of local token-level constraints versus global sentence-level objectives.

\textbf{Sentence-determined Advantage.} 
For sentence-level rewards, which depend on the entire response, we treat the sentence-level reward as a global feedback signal for the whole generated trajectory. For a given input $x$, the policy $\pi_\theta$ samples $G$ candidate sequences $\{y_1, \dots, y_G\}$ in one rollout, each associated with a sentence-level reward $r_i$ . 
Using intra-batch relative advantage, 
the sentence-level advantage is defined as
\begin{equation}
A_i^{\text{sentence}} = \frac{r_i - \text{mean}(r_1, \dots, r_G)}{\text{std}(r_1, \dots, r_G)}.
\end{equation}

Combining these two types of advantages, we define the final unified advantage at the token level for policy optimization
\begin{equation}
A_{i,t} = A_i^{\text{sentence}} + A_{i,t}^{\text{token}}.
\end{equation}
This formulation allows each token update to incorporate fine-grained local feedback from token-level constraints while simultaneously receiving global guidance from overall generation quality, achieving synergistic optimization of local behaviors and global objectives without introducing additional attribution assumptions.

\section{Experiment}
\label{sec:experiment}

\subsection{Experimental Setup}
In our experiments, we utilize \texttt{Qwen3-8B}~\cite{yang2025qwen3} as the base model for cold-start initialization, leveraging its strong semantic understanding to construct the initial policy distribution. 
All experiments are conducted on 8 NVIDIA A800 GPUs using ZeRO~\cite{20-zero} stage 2 from DeepSpeed~\cite{20-deepspeed}.
Regarding training hyperparameters, for each prompt we generate $5$ candidate responses online (i.e. $G=5$), enabling accurate group-relative advantage estimation. The global batch size per training step is set to $1440$, which ensures gradient stability while maintaining sufficient within-group sample diversity for comparison.

\subsubsection{Dataset}
In this work, we conduct experiments on a Baidu advertising dataset, which contains 400,000 real search ad samples covering user search intents (queries) and the ads’ core selling points. 
Search intents are explicit and immediate, requiring advertising texts to accurately reflect user intent while maintaining professional quality, making this dataset an ideal testbed for evaluating AI-generated high-quality creative content.

\subsubsection{Evaluation Metrics.}
To comprehensively evaluate our method in terms of both generation quality and real-world business impact, we adopt a multi-dimensional evaluation protocol that analyzes model performance from three complementary perspectives: business effectiveness, compliance and generation diversity.

\textbf{Business Effectiveness.}
We use $\Delta$CTCVR from online A/B testing under real production traffic, which measures the relative lift in CTCVR over the baseline.

It is defined as 
\begin{equation}
    \Delta \text{CTCVR} =
\frac{\text{CTCVR}_{\text{experiment}} - \text{CTCVR}_{\text{baseline}}}
{\text{CTCVR}_{\text{baseline}}}.
\end{equation}
This metric directly reflects the impact of the real-world business.

\textbf{Compliance.}
Compliance Rate quantifies the proportion of generated advertising texts that pass compliance and risk-control screening
\begin{equation}
    \text{Compliance Rate} =
\frac{1}{N} \sum_{i=1}^{N}
\mathbb{I}\big(y_i \in \mathcal{Y}_{\text{safe}}\big).
\end{equation}
$y_i$ denotes a generated advertising text. 
$\mathcal{Y}_{\text{safe}}$ is the set of texts approved by an industrial-grade risk control system deployed in a large-scale advertising platform, which performs automated compliance and risk assessment of advertising texts under advertising regulations.
$\mathbb{I}(\cdot)$is the indicator function. 
This metric ensures that performance gains are not achieved at the expense of regulatory compliance.

\textbf{Diversity Metric.}
To ensure consistency between optimization and evaluation, diversity is measured using the same reward formulation employed in training. 
Specifically, the diversity score is defined as
\begin{equation}
    \text{Diversity} =
\frac{1}{N} \sum_{i=1}^{N} r_i^{\text{div}}.
\end{equation}
$r_i^{\text{div}}$ denotes the diversity reward assigned to the $i$-th generated advertising text, capturing penalties for high-frequency patterns and duplicated expressions.

\subsubsection{Baselines}
To comprehensively evaluate the performance of our proposed RELATE method, we compare it with representative production and training-based baselines under real industrial settings:
\begin{itemize}
    \item \textbf{Historical Online Production Model (NLG).}
    A conventional pre-trained language model trained on high-CTR historical ad data, abbreviated as \texttt{NLG}.

    \item \textbf{Pre-RELATE Online Model (Qwen-SFT)} 
    , which fine-tunes \texttt{Qwen3-8B}~\cite{yang2025qwen3} via supervised learning on curated in-domain advertising data, including online posterior samples with high click signals, and applies post-processing at inference time to enforce compliance, formatting, and basic quality requirements. 
    It represents a commonly adopted industrial paradigm that combines supervised learning with non-end-to-end post-hoc constraints.
    
\end{itemize}

\begin{table}[h]
  \caption{
  Performance comparison of RELATE with other baseline methods.
  }
  \label{tab:main_results}
  \begin{tabular}{c|ccc}
    \toprule
    \textbf{Models} & \textbf{Compliance Rate}$\uparrow$ & \textbf{Diversity}$\uparrow$ & $\Delta$\textbf{CTCVR}$\uparrow$\\
    \midrule
    NLG & 73.00\% & 0.75 & -  \\
    Qwen-SFT & 82.70\% & 0.72 & +5.25\%  \\
    \textbf{RELATE} & \textbf{93.98\%} & \textbf{0.82} & \textbf{+9.19\%} \\
   \bottomrule
  \end{tabular}
\end{table}

\begin{table*}[t]
  \caption{
  Ablation configurations for reward design in Reinforcement Learning.
  }
  \label{tab:ablation_setup}
  \begin{tabular}{c|cccccc}
    \toprule
    \multirow{2}{*}{\textbf{Model ID}} & \textbf{Structural} & \multirow{2}{*}{\textbf{CTCVR}} & \multirow{2}{*}{\textbf{Diversity}} & \textbf{Semantic} & \textbf{Credit} & \textbf{Reward}\\
     & \textbf{Quality} &  &  & \textbf{Quality} & \textbf{Assignment} & \textbf{Granularity}\\
    \midrule
    Model 1 & \ding{51}  & \ding{55}  & \ding{55} & \ding{55} & \ding{55} & Sentence-level  \\
    Model 2 & \ding{51}  & \ding{51}  & \ding{55} & \ding{55} & \ding{55} & Sentence-level  \\
    Model 3 & \ding{51}  & \ding{51}  & \ding{51} & \ding{55} & \ding{55} & Sentence-level  \\
    Model 4 & \ding{51}  & \ding{51}  & \ding{51} & \ding{51} & \ding{55} & Sentence-level  \\
    RELATE & \ding{51}  & \ding{51}  & \ding{51} & \ding{51} & \ding{51} & Token-level (Diversity \& Risk Control)  \\
   \bottomrule
  \end{tabular}
\end{table*}

\subsection{Experimental Results}
\label{sec:exp_main_result}
The comparison focuses on three key dimensions that are critical for advertising text generation: business effectiveness, compliance and generation diversity.
Table~\ref{tab:main_results} reports the overall comparison results across different methods.
RELATE consistently outperforms both baselines across all evaluation dimensions. 
Compared with the production system, our method achieves substantial gains in compliance and diversity while delivering a significant improvement in online conversion performance. 
Compared with the supervised fine-tuning baseline with post-processing, RELATE yields markedly higher conversion lift and diversity, while further improving compliance performance.

\subsection{Human Evaluation}
To further assess the subjective quality of generated advertising texts in terms of semantic validity and native-style expression, we conduct a human evaluation using the Good / Same / Bad (GSB) pairwise comparison protocol. 
Annotators are presented with pairs of advertising texts generated by two different models under the same advertising scenario, without being informed of the model identities. 
Each pair is evaluated along two complementary aspects—semantic correctness and naturalness of expression—and a holistic judgment is made on which advertising text exhibits higher overall quality.
We assess our proposed RELATE against Qwen-SFT. 

\textbf{Evaluation results.}
Among the 300 comparison samples, RELATE shows a significant improvement compared to the baseline, with a GSB evaluation result of 49:42:9 (good : same : bad), indicating that RELATE is manually evaluated as better in nearly half of the samples and significantly reduced the proportion of bad samples.
This demonstrates that incorporating reinforcement learning with granularity-aware reward modeling enables the model to improve expressive richness and naturalness while maintaining semantic correctness.
Overall, the human evaluation results are well aligned with both offline metrics and online A/B testing outcomes, providing complementary evidence that {RELATE} achieves tangible improvements not only in quantitative business indicators but also in user-perceived generation quality.

\textbf{Case study.} 
We present several real-world online advertising texts in Table~\ref{tab:case_study} (displayed in Appendix~\ref{appendix:case_study} due to space limitation), which shows the differences between advertising texts generated by RELATE and those produced by baseline methods.
For each case, we disclose the baseline text and the RELATE-generated text displayed under the same search query and bid keyword.
These cases show that queries exhibit strong information-seeking intent, requiring advertising texts to not only cover essential factual explanations to ensure search relevance, but also to seamlessly integrate commercial service information under compliance constraints, thereby achieving a balance between search intent satisfaction and conversion-oriented persuasion.
From the table, we observe that baseline texts tend to respond to user queries with generic informational content. 
In contrast, while still fulfilling user search intent, RELATE more effectively organizes the content into advertising text expressions that align with advertising delivery logic.
These cases demonstrate that RELATE does not compromise responsiveness to user intent. 
Instead, through the synergistic effects of the CTCVR reward, diversity constraints, and semantic-level quality assessment via LLM-as-judge, RELATE guides the model toward a more favorable balance between information relevance and commercial conversion potential.
This observation is consistent with the conversion uplift trends observed in online A/B tests, validating the practical effectiveness of our approach in the context of search advertising text generation.



\begin{figure*}[!t]
\centering 
\captionsetup{aboveskip=0pt, belowskip=0pt} 
\includegraphics[width=1.0\linewidth,page=1]{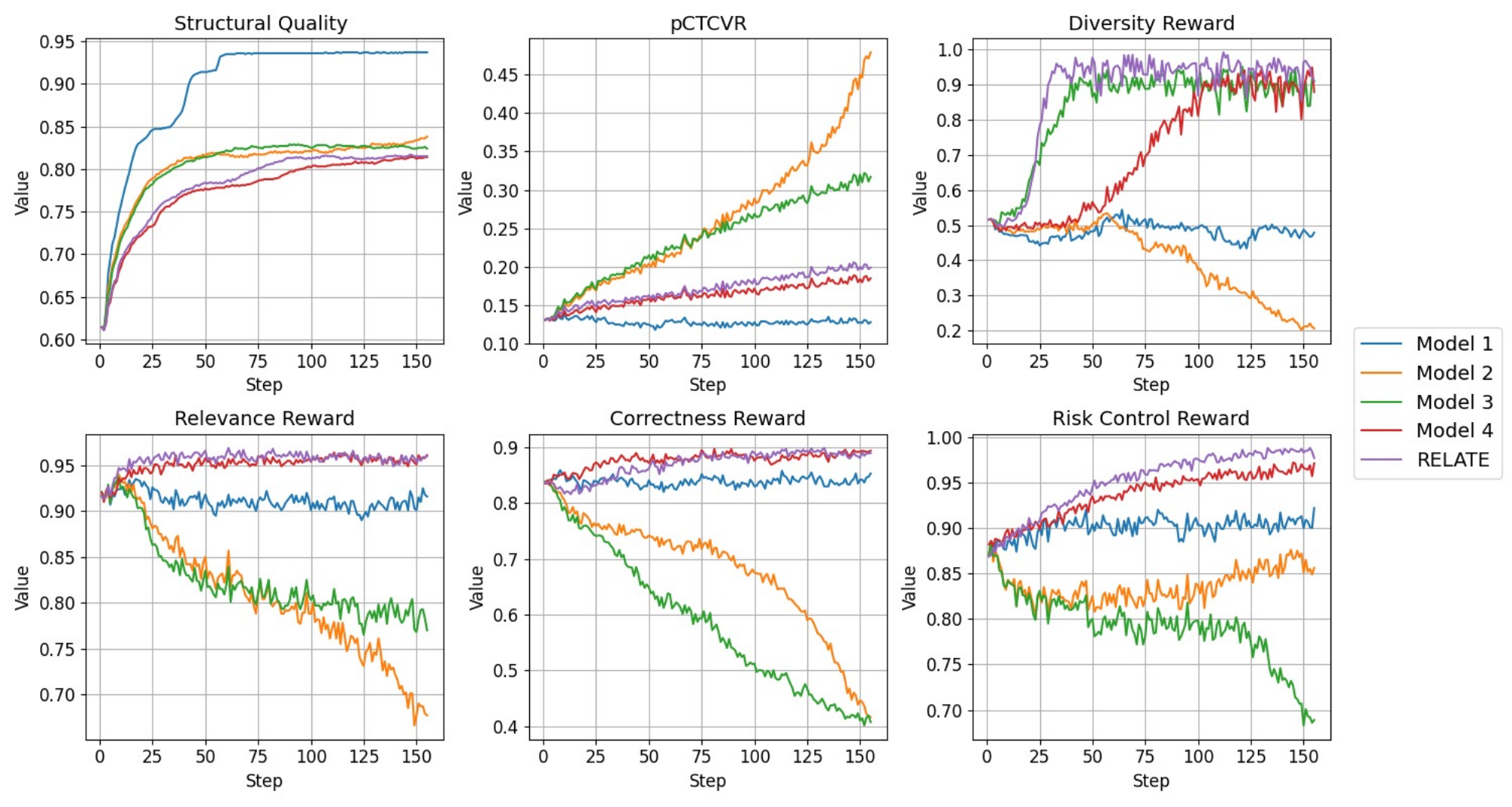}
\caption{
Training curves of individual rewards under different ablation settings.
}
\label{fig:ablation}
\end{figure*}

\subsection{Ablation Study}
\label{sec:ablation_study}
\subsubsection{Ablation Setup}
To systematically analyze the effects of different reward components in advertising text generation, we conduct a progressive ablation study centered on reward design. In reward modeling, the quality reward is treated as a unified objective and is further categorized into Structural Quality and Semantic Quality based on the nature of the constraints. 
Structural Quality consists of length and format rewards, which establish a minimal lower bound for usability and structural validity. Semantic Quality includes correctness, relevance, and risk-control rewards, capturing higher-level semantic alignment, factual reliability, and compliance requirements.

We construct five reward configurations, corresponding to Model 1–4 and RELATE in Table~\ref{tab:ablation_setup}, where each configuration introduces only one additional reward component on top of the previous setting, ensuring that performance variations can be clearly attributed. For instance, Model 1 implies only structural quality reward is included in GRPO training.Model 2, compared to Model 1, additionally incorporates CTCVR into the reward signal.Full reward configurations are summarized in Table~\ref{tab:ablation_setup}.

Beyond the progressive activation of quality dimensions, all reward signals in Models 1–4 are computed and backpropagated at the sentence level, serving as holistic evaluations of the final generated outputs. 
In contrast, RELATE keeps the reward set unchanged but introduces a Credit Assignment mechanism that attributes diversity- and semantic-quality-related constraints at the token level. 
This design allows subsequent ablation results to disentangle the effects of reward type expansion from those of credit assignment granularity.

\begin{table}[h]
  \caption{
  Performances of different configurations at step $150$.
  }
  \label{tab:ablation_results}
  \resizebox{\columnwidth}{!}{
  \begin{tabular}{c|cccc}
    \toprule
    \multirow{2}{*}{\textbf{Model ID}} & \textbf{Structural} & \multirow{2}{*}{\textbf{CTCVR(\%)}} & \multirow{2}{*}{\textbf{Diversity}} & \textbf{Semantic}\\
     & \textbf{Quality} & & & \textbf{Quality}\\
    \midrule
    Model 1 & 0.937  & 0.128  & 0.478 & 0.881 \\
    Model 2 & 0.834  & 0.450  & 0.211 & 0.658 \\
    Model 3 & 0.824  & 0.320  & 0.894 & 0.621 \\
    Model 4 & 0.812  & 0.184  & 0.923 & 0.935 \\
    RELATE & 0.815  & 0.203  & 0.965 & 0.939 \\
   \bottomrule
  \end{tabular}
  }
\end{table}

\subsubsection{Ablation Results}
Figure~\ref{fig:ablation} illustrates the dynamic evolution of various rewards during training across different ablation configurations. The curves reveal how individual objectives change over time and interact with one another under different reward settings. Complementing this, Table~\ref{tab:ablation_results} summarizes the model performances at step 150, providing a quantitative snapshot of Structural Quality, CTCVR, Diversity, and Semantic Quality for each configuration. Together, these visualizations capture both the training dynamics and the resulting trade-offs in final metrics.The ablation analysis reveals a systematic progression in model behavior across three phases:single-reward pursuit, multi-reward coordination, and token-level credit assignment.

Single-reward pursuit (Model 1–2): The blue model (Model 1) enforces only Structural Quality, achieving the highest structural scores but low CTCVR, Diversity, and Semantic Quality. When the CTCVR reward is added in the yellow model (Model 2), business metrics significantly improve, but Diversity and semantic scores decline. This indicates that while Structural Quality is maintained, the additional CTCVR reward drives the model to prioritize business objectives, negatively affecting diversity and semantic content. Under sentence-level reward attribution, the new business reward implicitly competes with other inactive quality constraints, leading to declines in some metrics.

Multi-reward coordination (Model 3–4): Adding a sentence-level Diversity reward in the green model (Model 3) partially alleviates mode collapse and significantly improves Diversity over the baseline. Relative to Models 1–2, however, the rate of CTCVR improvement slows, while Structural Quality remains largely stable. Incorporating Semantic Quality in the red model (Model 4) enhances correctness, relevance, and compliance, but Structural Quality and CTCVR improvement rate decelerates even more compared with Models 1–2. These observations demonstrate that sentence-level multi-reward optimization inherently introduces conflicts, as multiple objectives share gradients and compete, limiting exploration and diluting individual metric performance.

Token-level credit assignment: The purple model, RELATE, keeps the same reward set but introduces token-level credit assignment, enabling finer attribution of diversity and semantic quality signals.This mechanism improves Structural Quality, CTCVR, Diversity, and Semantic Quality. 

Overall, while the ablation analysis reveals trade-offs among different objectives in multi-reward optimization, the full RELATE system still outperforms the baseline in overall business metrics and semantic relevance, validating the effectiveness of the framework.

\section{Conclusions}
This paper introduces RELATE (Reinforcement-Learning-Enhanced LLM Framework for Advertising Text Generation), an end-to-end reinforcement learning framework for creative text generation in real-world advertising scenarios. 
RELATE builds upon large language models and systematically incorporates reward modeling to unify business constraint objectives—originally handled via post-processing or manual rules—directly into the generation process. 
These constraints include semantic relevance, factual and compliance correctness, diversity control, and final business conversion metrics (CTCVR). 
Deployed in an online advertising system, RELATE achieves a +9.19\% improvement in CTCVR. 
In the future, we will explore richer reward designs, such as semantic diversity rewards, and more fine-grained credit assignment strategies.


\bibliographystyle{ACM-Reference-Format}
\bibliography{ref}


\begin{thebibliography}{26}


\ifx \showCODEN    \undefined \def \showCODEN     #1{\unskip}     \fi
\ifx \showISBNx    \undefined \def \showISBNx     #1{\unskip}     \fi
\ifx \showISBNxiii \undefined \def \showISBNxiii  #1{\unskip}     \fi
\ifx \showISSN     \undefined \def \showISSN      #1{\unskip}     \fi
\ifx \showLCCN     \undefined \def \showLCCN      #1{\unskip}     \fi
\ifx \shownote     \undefined \def \shownote      #1{#1}          \fi
\ifx \showarticletitle \undefined \def \showarticletitle #1{#1}   \fi
\ifx \showURL      \undefined \def \showURL       {\relax}        \fi
\providecommand\bibfield[2]{#2}
\providecommand\bibinfo[2]{#2}
\providecommand\natexlab[1]{#1}
\providecommand\showeprint[2][]{arXiv:#2}

\bibitem[grp(2024)]%
        {grpolambda2024}
 \bibinfo{year}{2024}\natexlab{}.
\newblock \showarticletitle{GRPO-$\lambda$: Improving Credit Assignment in Reinforcement Learning for Large Language Models}.
\newblock \bibinfo{journal}{\emph{arXiv preprint arXiv:2510.00194}} (\bibinfo{year}{2024}).
\newblock


\bibitem[seg(2024)]%
        {segmentpo2024}
 \bibinfo{year}{2024}\natexlab{}.
\newblock \showarticletitle{Segment Policy Optimization: Effective Segment-Level Credit Assignment in RL for Large Language Models}.
\newblock \bibinfo{journal}{\emph{arXiv preprint arXiv:2505.23564}} (\bibinfo{year}{2024}).
\newblock


\bibitem[sto(2024)]%
        {stop_summation2024}
 \bibinfo{year}{2024}\natexlab{}.
\newblock \showarticletitle{Stop Summation: Min-Form Credit Assignment for Process Reward Models}.
\newblock \bibinfo{journal}{\emph{arXiv preprint arXiv:2504.15275}} (\bibinfo{year}{2024}).
\newblock


\bibitem[cap(2025)]%
        {capo2025}
 \bibinfo{year}{2025}\natexlab{}.
\newblock \showarticletitle{CAPO: Verifiable Generative Credit Assignment for LLM Reasoning}.
\newblock \bibinfo{journal}{\emph{arXiv preprint}} (\bibinfo{year}{2025}).
\newblock


\bibitem[Bartz et~al\mbox{.}(2008)]%
        {bartz2008natural}
\bibfield{author}{\bibinfo{person}{Kevin Bartz}, \bibinfo{person}{Cory Barr}, {and} \bibinfo{person}{Adil Aijaz}.} \bibinfo{year}{2008}\natexlab{}.
\newblock \showarticletitle{Natural language generation for sponsored-search advertisements}. In \bibinfo{booktitle}{\emph{Proceedings of the 9th ACM Conference on Electronic Commerce}}. \bibinfo{pages}{1--9}.
\newblock


\bibitem[Cao et~al\mbox{.}(2025)]%
        {cao2025perspective}
\bibfield{author}{\bibinfo{person}{Yilin Cao}, \bibinfo{person}{Ruike Zhang}, \bibinfo{person}{Penghui Wei}, \bibinfo{person}{Qingchao Kong}, {and} \bibinfo{person}{Wenji Mao}.} \bibinfo{year}{2025}\natexlab{}.
\newblock \showarticletitle{Perspective-driven Preference Optimization with Entropy Maximization for Diverse Argument Generation}. In \bibinfo{booktitle}{\emph{Findings of the Association for Computational Linguistics: EMNLP 2025}}. \bibinfo{pages}{22479--22496}.
\newblock


\bibitem[Chen et~al\mbox{.}(2025)]%
        {chen2025ctr}
\bibfield{author}{\bibinfo{person}{Yanda Chen}, \bibinfo{person}{Zihui Ren}, \bibinfo{person}{Qixiang Gao}, \bibinfo{person}{Jiale Chen}, \bibinfo{person}{Si Chen}, \bibinfo{person}{Xubin Li}, \bibinfo{person}{Tiezheng Ge}, {and} \bibinfo{person}{Bo Zheng}.} \bibinfo{year}{2025}\natexlab{}.
\newblock \showarticletitle{CTR-Driven Ad Text Generation via Online Feedback Preference Optimization}.
\newblock \bibinfo{journal}{\emph{arXiv preprint arXiv:2507.20227}} (\bibinfo{year}{2025}).
\newblock


\bibitem[Hughes et~al\mbox{.}(2019)]%
        {hughes2019generating}
\bibfield{author}{\bibinfo{person}{J~Weston Hughes}, \bibinfo{person}{Keng-hao Chang}, {and} \bibinfo{person}{Ruofei Zhang}.} \bibinfo{year}{2019}\natexlab{}.
\newblock \showarticletitle{Generating better search engine text advertisements with deep reinforcement learning}. In \bibinfo{booktitle}{\emph{Proceedings of the 25th ACM SIGKDD international conference on knowledge discovery \& data mining}}. \bibinfo{pages}{2269--2277}.
\newblock


\bibitem[Jin et~al\mbox{.}(2023)]%
        {jin2023towards}
\bibfield{author}{\bibinfo{person}{Yiping Jin}, \bibinfo{person}{Akshay Bhatia}, \bibinfo{person}{Dittaya Wanvarie}, {and} \bibinfo{person}{Phu~TV Le}.} \bibinfo{year}{2023}\natexlab{}.
\newblock \showarticletitle{Towards improving coherence and diversity of slogan generation}.
\newblock \bibinfo{journal}{\emph{Natural Language Engineering}} \bibinfo{volume}{29}, \bibinfo{number}{2} (\bibinfo{year}{2023}), \bibinfo{pages}{254--286}.
\newblock


\bibitem[Ma et~al\mbox{.}(2018)]%
        {ma2018entire}
\bibfield{author}{\bibinfo{person}{Xiao Ma}, \bibinfo{person}{Liqin Zhao}, \bibinfo{person}{Guan Huang}, \bibinfo{person}{Zhi Wang}, \bibinfo{person}{Xiaoqiang Hu}, \bibinfo{person}{Zhihong Zhang}, {and} \bibinfo{person}{Kun Gai}.} \bibinfo{year}{2018}\natexlab{}.
\newblock \showarticletitle{Entire Space Multi-Task Model: An Effective Approach for Estimating Post-Click Conversion Rate}. In \bibinfo{booktitle}{\emph{Proceedings of the 24th ACM SIGKDD International Conference on Knowledge Discovery \& Data Mining}}. \bibinfo{publisher}{ACM}, \bibinfo{pages}{1137--1146}.
\newblock


\bibitem[Murakami et~al\mbox{.}(2023)]%
        {nlg_ad_survey2026}
\bibfield{author}{\bibinfo{person}{Soichiro Murakami}, \bibinfo{person}{Sho Hoshino}, {and} \bibinfo{person}{Peinan Zhang}.} \bibinfo{year}{2023}\natexlab{}.
\newblock \showarticletitle{Natural language generation for advertising: A survey}.
\newblock \bibinfo{journal}{\emph{arXiv preprint arXiv:2306.12719}} (\bibinfo{year}{2023}).
\newblock


\bibitem[Rafailov et~al\mbox{.}(2023)]%
        {rafailov2023dpo}
\bibfield{author}{\bibinfo{person}{Rafael Rafailov}, \bibinfo{person}{Archit Sharma}, \bibinfo{person}{Eric Mitchell}, {et~al\mbox{.}}} \bibinfo{year}{2023}\natexlab{}.
\newblock \showarticletitle{Direct Preference Optimization: Your Language Model is Secretly a Reward Model}.
\newblock \bibinfo{journal}{\emph{arXiv preprint arXiv:2305.18290}} (\bibinfo{year}{2023}).
\newblock


\bibitem[Rajbhandari et~al\mbox{.}(2020)]%
        {20-zero}
\bibfield{author}{\bibinfo{person}{Samyam Rajbhandari}, \bibinfo{person}{Jeff Rasley}, \bibinfo{person}{Olatunji Ruwase}, {and} \bibinfo{person}{Yuxiong He}.} \bibinfo{year}{2020}\natexlab{}.
\newblock \showarticletitle{ZeRO: memory optimizations toward training trillion parameter models}. In \bibinfo{booktitle}{\emph{SC}}. \bibinfo{pages}{1--16}.
\newblock


\bibitem[Rasley et~al\mbox{.}(2020)]%
        {20-deepspeed}
\bibfield{author}{\bibinfo{person}{Jeff Rasley}, \bibinfo{person}{Samyam Rajbhandari}, \bibinfo{person}{Olatunji Ruwase}, {and} \bibinfo{person}{Yuxiong He}.} \bibinfo{year}{2020}\natexlab{}.
\newblock \showarticletitle{Deepspeed: System optimizations enable training deep learning models with over 100 billion parameters}. In \bibinfo{booktitle}{\emph{KDD}}. \bibinfo{pages}{3505--3506}.
\newblock


\bibitem[Shao et~al\mbox{.}(2024)]%
        {shao2024grpo}
\bibfield{author}{\bibinfo{person}{Zhihong Shao}, \bibinfo{person}{Peiyi Wang}, \bibinfo{person}{Qihao Zhu}, \bibinfo{person}{Runxin Xu}, \bibinfo{person}{Junxiao Song}, \bibinfo{person}{Xiao Bi}, \bibinfo{person}{Haowei Zhang}, \bibinfo{person}{Mingchuan Zhang}, \bibinfo{person}{YK Li}, \bibinfo{person}{Yang Wu}, {et~al\mbox{.}}} \bibinfo{year}{2024}\natexlab{}.
\newblock \showarticletitle{Deepseekmath: Pushing the limits of mathematical reasoning in open language models}.
\newblock \bibinfo{journal}{\emph{arXiv preprint arXiv:2402.03300}} (\bibinfo{year}{2024}).
\newblock


\bibitem[Thomaidou et~al\mbox{.}(2013)]%
        {thomaidou2013grammads}
\bibfield{author}{\bibinfo{person}{Stamatina Thomaidou}, \bibinfo{person}{Konstantinos Leymonis}, {and} \bibinfo{person}{Michalis Vazirgiannis}.} \bibinfo{year}{2013}\natexlab{}.
\newblock \showarticletitle{GrammAds: Keyword and ad creative generator for online advertising campaigns}. In \bibinfo{booktitle}{\emph{Digital Enterprise Design and Management 2013: Proceedings of the First International Conference on Digital Enterprise Design and Management DED\&M 2013}}. Springer, \bibinfo{pages}{33--44}.
\newblock


\bibitem[Vasudevan et~al\mbox{.}(2025)]%
        {vasudevan2025llm}
\bibfield{author}{\bibinfo{person}{Varun Vasudevan}, \bibinfo{person}{Faezeh Akhavizadegan}, \bibinfo{person}{Abhinav Prakash}, \bibinfo{person}{Yokila Arora}, \bibinfo{person}{Jason Cho}, \bibinfo{person}{Tanya Mendiratta}, \bibinfo{person}{Sushant Kumar}, {and} \bibinfo{person}{Kannan Achan}.} \bibinfo{year}{2025}\natexlab{}.
\newblock \showarticletitle{LLM-driven Constrained Copy Generation through Iterative Refinement}.
\newblock \bibinfo{journal}{\emph{arXiv preprint arXiv:2504.10391}} (\bibinfo{year}{2025}).
\newblock


\bibitem[Wang et~al\mbox{.}(2021)]%
        {wang2021reinforcing}
\bibfield{author}{\bibinfo{person}{Xiting Wang}, \bibinfo{person}{Xinwei Gu}, \bibinfo{person}{Jie Cao}, \bibinfo{person}{Zihua Zhao}, \bibinfo{person}{Yulan Yan}, \bibinfo{person}{Bhuvan Middha}, {and} \bibinfo{person}{Xing Xie}.} \bibinfo{year}{2021}\natexlab{}.
\newblock \showarticletitle{Reinforcing pretrained models for generating attractive text advertisements}. In \bibinfo{booktitle}{\emph{Proceedings of the 27th ACM SIGKDD conference on knowledge discovery \& data mining}}. \bibinfo{pages}{3697--3707}.
\newblock


\bibitem[Wei et~al\mbox{.}(2022)]%
        {wei2022creater}
\bibfield{author}{\bibinfo{person}{Penghui Wei}, \bibinfo{person}{Xuanhua Yang}, \bibinfo{person}{Shaoguo Liu}, \bibinfo{person}{Liang Wang}, {and} \bibinfo{person}{Bo Zheng}.} \bibinfo{year}{2022}\natexlab{}.
\newblock \showarticletitle{CREATER: CTR-driven advertising text generation with controlled pre-training and contrastive fine-tuning}.
\newblock \bibinfo{journal}{\emph{arXiv preprint arXiv:2205.08943}} (\bibinfo{year}{2022}).
\newblock


\bibitem[Yang et~al\mbox{.}(2025)]%
        {yang2025qwen3}
\bibfield{author}{\bibinfo{person}{An Yang}, \bibinfo{person}{Anfeng Li}, \bibinfo{person}{Baosong Yang}, \bibinfo{person}{Beichen Zhang}, \bibinfo{person}{Binyuan Hui}, \bibinfo{person}{Bo Zheng}, \bibinfo{person}{Bowen Yu}, \bibinfo{person}{Chang Gao}, \bibinfo{person}{Chengen Huang}, \bibinfo{person}{Chenxu Lv}, {et~al\mbox{.}}} \bibinfo{year}{2025}\natexlab{}.
\newblock \showarticletitle{Qwen3 technical report}.
\newblock \bibinfo{journal}{\emph{arXiv preprint arXiv:2505.09388}} (\bibinfo{year}{2025}).
\newblock


\bibitem[Youngmann et~al\mbox{.}(2021)]%
        {youngmann2021algorithmic}
\bibfield{author}{\bibinfo{person}{Brit Youngmann}, \bibinfo{person}{Elad Yom-Tov}, \bibinfo{person}{Ran Gilad-Bachrach}, {and} \bibinfo{person}{Danny Karmon}.} \bibinfo{year}{2021}\natexlab{}.
\newblock \showarticletitle{Algorithmic copywriting: Automated generation of health-related advertisements to improve their performance}.
\newblock \bibinfo{journal}{\emph{Information retrieval journal}} \bibinfo{volume}{24}, \bibinfo{number}{3} (\bibinfo{year}{2021}), \bibinfo{pages}{205--239}.
\newblock


\bibitem[Zhang et~al\mbox{.}(2021)]%
        {zhang2021chase}
\bibfield{author}{\bibinfo{person}{Chao Zhang}, \bibinfo{person}{Jingbo Zhou}, \bibinfo{person}{Xiaoling Zang}, \bibinfo{person}{Qing Xu}, \bibinfo{person}{Liang Yin}, \bibinfo{person}{Xiang He}, \bibinfo{person}{Lin Liu}, \bibinfo{person}{Haoyi Xiong}, {and} \bibinfo{person}{Dejing Dou}.} \bibinfo{year}{2021}\natexlab{}.
\newblock \showarticletitle{CHASE: Commonsense-enriched advertising on search engine with explicit knowledge}. In \bibinfo{booktitle}{\emph{Proceedings of the 30th ACM International Conference on Information \& Knowledge Management}}. \bibinfo{pages}{4352--4361}.
\newblock


\bibitem[Zhang et~al\mbox{.}(2023)]%
        {zhang2023planner}
\bibfield{author}{\bibinfo{person}{Yizhe Zhang}, \bibinfo{person}{Jiatao Gu}, \bibinfo{person}{Zhuofeng Wu}, \bibinfo{person}{Shuangfei Zhai}, \bibinfo{person}{Joshua Susskind}, {and} \bibinfo{person}{Navdeep Jaitly}.} \bibinfo{year}{2023}\natexlab{}.
\newblock \showarticletitle{Planner: Generating diversified paragraph via latent language diffusion model}.
\newblock \bibinfo{journal}{\emph{Advances in Neural Information Processing Systems}}  \bibinfo{volume}{36} (\bibinfo{year}{2023}), \bibinfo{pages}{80178--80190}.
\newblock


\bibitem[Zhang et~al\mbox{.}(2025)]%
        {zhang2025qwen3embedding}
\bibfield{author}{\bibinfo{person}{Yanzhao Zhang}, \bibinfo{person}{Mingxin Li}, \bibinfo{person}{Dingkun Long}, \bibinfo{person}{Xin Zhang}, \bibinfo{person}{Huan Lin}, \bibinfo{person}{Baosong Yang}, \bibinfo{person}{Pengjun Xie}, \bibinfo{person}{An Yang}, \bibinfo{person}{Dayiheng Liu}, \bibinfo{person}{Junyang Lin}, {et~al\mbox{.}}} \bibinfo{year}{2025}\natexlab{}.
\newblock \showarticletitle{Qwen3 Embedding: Advancing Text Embedding and Reranking Through Foundation Models}.
\newblock \bibinfo{journal}{\emph{arXiv preprint arXiv:2506.05176}} (\bibinfo{year}{2025}).
\newblock


\bibitem[Zhang et~al\mbox{.}(2024)]%
        {zhang2024adtec}
\bibfield{author}{\bibinfo{person}{Yifan Zhang}, \bibinfo{person}{Hao Liu}, \bibinfo{person}{Jianan Wang}, {et~al\mbox{.}}} \bibinfo{year}{2024}\natexlab{}.
\newblock \showarticletitle{AdTEC: A Unified Benchmark for Evaluating Text Quality in Search Engine Advertising}.
\newblock \bibinfo{journal}{\emph{arXiv preprint arXiv:2408.05906}} (\bibinfo{year}{2024}).
\newblock


\bibitem[Zhou et~al\mbox{.}(2023)]%
        {zhou2023creater}
\bibfield{author}{\bibinfo{person}{Yujia Zhou}, \bibinfo{person}{Wei Li}, \bibinfo{person}{Yifan Zhang}, {et~al\mbox{.}}} \bibinfo{year}{2023}\natexlab{}.
\newblock \showarticletitle{CREATER: CTR-Driven Advertising Text Generation}.
\newblock \bibinfo{journal}{\emph{arXiv preprint arXiv:2310.09234}} (\bibinfo{year}{2023}).
\newblock


\end{thebibliography}

\appendix

\newpage
\FloatBarrier
\onecolumn
\section{Online Advertising Texts}
\label{appendix:case_study}
{
\renewcommand{\arraystretch}{1.0}
\begin{table*}[b]
  \caption{
  Advertising texts comparison. 
  Three advertising texts are generated by baselines and RELATE respectively.
  The actual advertising text is in Chinese, and its English translation is displayed below the Chinese content.
  }
  \label{tab:case_study}
  \begin{tabular}{M{2.2cm}|M{1.5cm}|M{1.5cm}|L{5.5cm}|L{5.5cm}}
    \toprule
    \textbf{Search Query} 
    & \textbf{Bidword} 
    & \textbf{Text Type} 
    & \multicolumn{1}{c|}{\textbf{Baselines}} 
    & \multicolumn{1}{c}{\textbf{RELATE}} \\
    \midrule
    \multirow{4}{*}{
      \parbox[c][6.0cm][c]{2.0cm}{
        \centering
        \begin{CJK}{UTF8}{gbsn}
        最有效的\\减肥方法
        \end{CJK} \\
        most effective weight loss method
      }}
    &
    \multirow{4}{*}{
      \parbox[c][6.0cm][c]{1.5cm}{
        \centering
        \begin{CJK}{UTF8}{gbsn}
        食疗减肥
        \end{CJK} \\
        dietary therapy for weight loss
      }}
    &
    \multirow{2}{*}{\parbox[c][1.1cm][c]{1.0cm}{\centering Title}}
    &
    \begin{CJK}{UTF8}{gbsn}
    \small{减肥食谱一周七天一日三餐-营养均衡}
    \end{CJK} 
    &
    \begin{CJK}{UTF8}{gbsn}
    \small{科学减重分阶段，效果稳也更健康}
    \end{CJK} \\
    & & & 
    \textit{7-Day Weight Loss Meal Plan with Balanced Nutrition (Three Meals a Day)}
    &
    \textit{Stage-Based, Science-Backed Weight Loss for Steadier and Healthier Results}
    \\
    \cline{3-5}
    & 
    &
    \multirow{2}{*}{\parbox[c][5.2cm][c]{1.5cm}{\centering Description}}
    &
    \vspace{0.01cm}
    \begin{CJK}{UTF8}{gbsn}
    \small{合理的一日三餐帮你建立健康代谢。
    90天是脂肪细胞更新周期，不追求快速瘦身，三个月养成易瘦体质，效果更持久。}
    \end{CJK} 
    &
    \vspace{0.01cm}
    \begin{CJK}{UTF8}{gbsn}
    \small{减肥想要效果快又健康，应分阶段进行，不能盲目追求速度。初期体重下降0.5-1kg/周，
    热量摄入每日1200-1500大卡，搭配高蛋白、低升糖食物。长期结合运动维持代谢。}
    \end{CJK} \\
    & 
    &
    &
    \parbox[c][2.8cm][c]{5.5cm}{
    \small{A well-planned daily three-meal structure helps build a healthy metabolism. Ninety days correspond to the renewal cycle of fat cells; instead of pursuing rapid weight loss, focusing on three months of consistent, balanced eating helps develop a body that is naturally easier to stay lean, with more sustainable results.}
    } 
    &
    \parbox[c][3.0cm][c]{5.5cm}{
    \small{Effective and healthy weight loss should be approached in stages rather than by blindly chasing speed. In the initial phase, a gradual loss of 0.5–1 kg per week is recommended, with a daily calorie intake of 1,200–1,500 kcal, combined with high-protein and low-glycemic foods. Over the long term, incorporating regular exercise helps maintain metabolism and sustain results.}
    }
    \\
    \midrule
    \multirow{4}{*}{
      \parbox[c][6.3cm][c]{2.0cm}{
        \centering
        \begin{CJK}{UTF8}{gbsn}
        近视怎么矫正
        \end{CJK} \\
        how to correct myopia
      }}
    &
    \multirow{4}{*}{
      \parbox[c][6.3cm][c]{1.5cm}{
        \centering
        \begin{CJK}{UTF8}{gbsn}
        近视
        \end{CJK} \\
        myopia
      }}
    &
    \multirow{2}{*}{\parbox[c][1.1cm][c]{1.0cm}{\centering Title}}
    &
    \begin{CJK}{UTF8}{gbsn}
    \small{5-15岁孩子近视怎么办? 宝妈要重视}
    \end{CJK} 
    &
    \begin{CJK}{UTF8}{gbsn}
    \small{孩子近视要警惕！抓住视力矫正黄金期}
    \end{CJK} \\
    & & & 
    \textit{What Should You Do If a Child Aged 5–15 Has Myopia? Moms Should Take It Seriously}
    &
    \textit{Stay Alert to Childhood Myopia! Seize the Critical Window for Vision Correction}
    \\
    \cline{3-5}
    & 
    &
    \multirow{2}{*}{\parbox[c][4.7cm][c]{1.5cm}{\centering Description}}
    &
    \vspace{0.01cm}
    \begin{CJK}{UTF8}{gbsn}
    \small{孩子近视能调整吗？孩子不良的用眼习惯，导致视力不断下降，眼轴不断增长，家长不必过于内疚，在家尝试一下这个方法，每天15分钟，控制度数增长。}
    \end{CJK} 
    &
    \vspace{0.01cm}
    \begin{CJK}{UTF8}{gbsn}
    \small{孩子近视不是只能“戴镜到老”。对于轻中度近视，有干预空间但要把握黄金期。低度近视规范用眼管理，中度近视实施个性化光学矫正，延缓近视进展。}
    \end{CJK} \\
    & 
    &
    &
    \parbox[c][2.8cm][c]{5.5cm}{
    \small{Can childhood myopia be adjusted? Poor eye-use habits can cause vision to decline and the eyeball length to increase. Parents don’t need to feel overly guilty—try this method at home for just 15 minutes a day to help control the progression of myopia.}
    }
    &
    \parbox[c][3.0cm][c]{5.5cm}{
    \small{Childhood myopia does not mean a lifetime of wearing glasses. For mild to moderate myopia, effective intervention is possible—but only if the critical window is properly utilized. Mild myopia can be managed through standardized eye-use habits, while moderate myopia requires individualized optical correction to slow the progression of myopia.}
    }
    \\
    \midrule
    \multirow{4}{*}{
      \parbox[c][6.3cm][c]{2.0cm}{
        \centering
        \begin{CJK}{UTF8}{gbsn}
        脂溢性脱发\\怎么治
        \end{CJK} \\
        how to treat seborrheic alopecia
      }}
    &
    \multirow{4}{*}{
      \parbox[c][6.3cm][c]{1.5cm}{
        \centering
        \begin{CJK}{UTF8}{gbsn}
        脂溢性\\脱发
        \end{CJK} \\
        seborrheic alopecia
      }}
    &
    \multirow{2}{*}{\parbox[c][1.1cm][c]{1.0cm}{\centering Title}}
    &
    \begin{CJK}{UTF8}{gbsn}
    \small{脂溢性脱发是怎么引起的，有什么好办法吗}
    \end{CJK} 
    &
    \begin{CJK}{UTF8}{gbsn}
    \small{脂溢性脱发怎么办，科学治疗有希望}
    \end{CJK} \\
    & & & 
    \textit{What Causes Seborrheic Alopecia? Are There Effective Solutions?}
    &
    \textit{What to Do About Seborrheic Alopecia? Science-Based Treatment Brings Hope}
    \\
    \cline{3-5}
    & 
    &
    \multirow{2}{*}{\parbox[c][4.7cm][c]{1.5cm}{\centering Description}}
    &
    \vspace{0.01cm}
    \begin{CJK}{UTF8}{gbsn}
    \small{脂溢性脱发的原因，多为遗传，精神压力，气血衰弱原因，轻度脱发可规范作息，饮食均衡改善脱发症状，严重脱发患者可在线咨询脱发医生。}
    \end{CJK} 
    &
    \vspace{0.01cm}
    \begin{CJK}{UTF8}{gbsn}
    \small{脂溢性脱发与头皮油脂失衡有关，别着急，先弄清楚自己属于哪种情况。出油多脱发少: 注意温和清洁。出油重脱发多: 调整作息，少糖少脂。脱发反复持续：咨询皮肤科评估。}
    \end{CJK} \\
    & 
    &
    &
    \parbox[c][2.8cm][c]{5.5cm}{
    \small{Seborrheic alopecia is commonly caused by genetic factors, psychological stress, and weakened physical condition. Mild hair loss may improve with a regular routine and a balanced diet. For more severe cases, online consultation with a hair loss specialist is recommended.}
    } 
    &
    \parbox[c][3.5cm][c]{5.5cm}{
    \small{Seborrheic alopecia is closely linked to an imbalance in scalp oil production. There’s no need to panic. First, identify which pattern applies. If oiliness is high but hair loss is mild, focus on gentle cleansing. If both oiliness and shedding are pronounced, adjust daily routines and reduce sugar and fat intake. If hair loss is recurrent or persistent, a dermatology consultation is recommended for professional evaluation.}
    }
    \\
   \bottomrule
  \end{tabular}
\end{table*}
}

\end{document}